\documentclass{article}
\usepackage{ijcai19}

\usepackage{times}
\usepackage{soul}
\usepackage{url}
\usepackage[hidelinks]{hyperref}
\usepackage[utf8]{inputenc}
\usepackage[small]{caption}
\usepackage{graphicx}
\usepackage{amsmath}
\usepackage{booktabs}
\usepackage{algorithm}
\usepackage{algorithmic}
\usepackage{times}
\usepackage{epsfig}
\usepackage{amssymb}
\usepackage{shortbold}
\usepackage{amsfonts}       
\usepackage{nicefrac}       
\usepackage{microtype} 

\title{Bayesian Optimized Continual Learning with Attention Mechanism}

\author{
Ju Xu$^1$
\and
Jin Ma$^2$\and
Zhanxing Zhu$^3$
\affiliations
$^{1,2}$Center for Data Science, Peking University\\
$^3$Center for Data Science, Peking University\&Beijing Institute of Big Data Research \\
\emails
\{xuju, ma\_jin, zhanxing.zhu\}@pku.edu.cn
}

\begin{document}

\maketitle

\begin{abstract}
Though neural networks have achieved much progress in various applications, it is still highly challenging for them to learn from a continuous stream of tasks without forgetting. Continual learning, a new learning paradigm, aims to solve this issue. In this work, we propose a new model for continual learning, called Bayesian Optimized Continual Learning with Attention Mechanism (BOCL) that dynamically expands the network capacity upon the arrival of new tasks by Bayesian optimization and selectively utilizes previous knowledge (e.g. feature maps of previous tasks) via attention mechanism. Our experiments on variants of MNIST and CIFAR-100 demonstrate that our methods outperform the state-of-the-art in preventing catastrophic forgetting and fitting new tasks better.
\end{abstract}

\section{Introduction}
Continual learning~\cite{thrun1}, aiming to solve new tasks quicker without forgetting previous tasks, is a long-standing challenge for the development of artificial intelligence systems. Compared with transfer learning, continual learning not only learns new tasks well by leveraging knowledge from earlier tasks, but also should prevent forgetting previous tasks after learning new tasks.

While there are many different methods to achieve continual learning, we only consider continual learning in the scenario of deep learning.  There are two groups of methods attempting to alleviate the problem of forgetting previous tasks. One  is to keep the network architecture fixed and finetune parameters while learning new tasks. The other group is to expand the network architecture while new tasks come along. We will introduce the two groups below, respectively. 

The first category of methods maintain a fixed network architecture with large capacity. When a new task arrives, the model parameters will be finetuned and some regularization term will be apply to prevent forgetting previous tasks~\cite{kirkpatrick1,zenke1}. In~\cite{lee2017overcoming}, the authors proposed incremental moment matching (IMM) which incrementally matched the moment of the posterior distribution of the neural network to resolve the catastrophic forgetting problem. He and Jaeger~\cite{he2018overcoming} introduced conceptor-aided backpropagation, which used conceptors to shield gradients from forgetting prior knowledge. Ronald and Christopher~\cite{kemker2017fearnet} proposed to mitigate catastrophic forgetting by  employing a
generative autoencoder to generate previously learned examples that are replayed alongside novel
information during consolidation. A recent method named P\&C~\cite{schwarz2018progress} consists of a knowledge base and an active column,
which were trained in two distinct, alternating phases. During
the progress phase, only parameters in the active column were trained with the use of the knowledge base. After the completion of the progress phase, the active
column is distilled into the knowledge base via  Elastic Weight Consolidation~\cite{kirkpatrick1} to mitigate forgetting in the knowledge
base, thus forming
the compress phase. In ~\cite{serra2018overcoming}, the authors employed a hard attention (HAT) mask  learned concurrently to every task through stochastic gradient descent to achieve continual learning.  Lopez-Paz and Ranzato~\cite{GradientEpisodicMemory} proposed Gradient Episodic Memory (GEM) that stores some subsets of examples of previous tasks and trains the new task with these subsets. However, these approaches comprise additional loss terms for preventing catastrophic forgetting, and thus,  with a limited amount of neural resources, they will inevitably lead to a tradeoff on the performance of old and new tasks.

The other group of approaches for preventing catastrophic forgetting is to change network architecture in response to the new tasks. For instance, Rusu \textit{et al.}~\cite{rusu1} proposed progressive neural networks (PGN) to block any changes to the previous parameters and expand the architecture by instantiating
a pre-specified neural network for the new task. In PGN, previous learnt features were utilized via lateral connections.
However, PGN typically results in an extremely large network when facing with many tasks since the newly added components are never optimized. Such a large network is expensive to store and even unnecessary due to its high redundancy. Dynamically Expandable Network (DEN)~\cite{yoon1} considered group Lasso regularization to prune redundant parameters after expanding the network with a fixed size of nodes/filters and perform selective retraining over previous parameters. However, DEN is complicated due to that it contains a chain of sub-algorithms and has many hyperparameters.
Besides, DEN is difficult to prevent catastrophic forgetting completely since it conducts selective retraining. Xu and Zhu~\cite{xu2018reinforced} introduced Reinforced Continual Learning (RCL), which expands the architectures by reinforcement learning and strikes a balance between model complexity and model performance based on a sophisticated designed reward. However, RCL requires a large number of trials to reach satisfying performance. 
DEN and RCL utilize the previous learned knowledge directly,  however, which might deteriorate the performance of learning the new tasks. This is because some of the previous learned knowledge might be irrelevant and interfere with learning the new tasks. Though PGN utilizes the previous learned knowledge via lateral connections, the inefficiency might rise since lateral connections will introduce many additional parameters (e.g., projection matrix and lateral connections).   

In this work, we propose a new framework for continual learning to expand the network more sparsely and utilize previous knowledge better. Faced with a new task, deciding optimal number of nodes/filters to add for each layer is posed as a combinatorial optimization problem. 
Inspired by Bayesian optimization (BO) for tuning hyperparameters~\cite{bergstra2011algorithms,snoek2012practical}, we utilize it to determine the number of nodes/filters added for each layer. 
In order to speed up the search process, we will initialize the  search points based on previous tasks to warmstart the Gaussian process model. Inspired by attention mechanism in image caption~\cite{chen2017sca}, we employ attention mechanism to learn how important the previous knowledge is for the new task. To the best of our knowledge, the proposal is the first attempt that applies the Bayesian optimization and attention mechanism for solving the continual learning problems, which could strike a significantly better balance between performance, network complexity and training time than existing approaches.


\section{Preliminaries }
\label{sec:pre}
\subsection{Introduction to Bayesian optimization}
Suppose our objective function is $f(z)$, we want to maximize it on domain $\mathcal{Z}$, $z^* = \argmax_{z \in \mathcal{Z}} f(z)$.

If we know the objective function $f$ and it is differentiable, we can always utilize gradient descent to optimize it. However, if an exact functional form for $f$ is not available (i.e., $f$ becomes a ``black box'' function), then how to optimize it?
Bayesian optimization \cite{jones1998efficient} solves it by 
maintaining a probabilistic belief about $f$ and utilizing a acquisition function to determine where to evaluate the function next. 

We denote $z_i$ as the $i$-th sample, and $f(z_i)$ as the observations of the objective function at $z_i$. The accumulated observations are $\Dcal_{1:N} = \{ z_{i},f(z_{i}) \}_{i=1}^N$. We select the Gaussian process (GP) as the prior distribution of $f$, due to its flexibility and tractability. The prior distribution is combined with the likelihood function $\Pcal(\Dcal_{1:N}|f)$. Now, we can combine these to obtain
our posterior distribution over the ``black box'' function $f$: $\Pcal(f|\Dcal_{1:N}) \propto \Pcal(\Dcal_{1:N}|f)\Pcal(f) .$

Bayesian optimization utilizes an acquisition function to determine the next point $z_{N+1} \in \Zcal$. 
We optimize the acquisition function to select the location of the next observation.  In our work, we select the most popular one expected improvement (EI) \cite{jones1998efficient} as the acquisition function. We provide a brief introduction to EI. 


Modeled with a Gaussian process, the function value at a given point $z$ can be considered as a normal random variable with mean $\mu$ and variance $\sigma^2$. Suppose $f^*$ is the minimal value of $f$ observed so far. The improvement at $z$ corresponds to the  utility function $\Ical(z) = max(0, f(z) - f^*)$, 
where $\Ical(z)$ is a random variable. 
We can consider the EI to assess $z$, $E[\Ical(z)] = E_{Y \sim N(\mu,\sigma^2)}[\Ical(z)]$.
With the reparameterization trick, $Y = \mu + \sigma \epsilon$ where $\epsilon \sim \Ncal(0,1)$, and we have $E[\Ical(z)] = E_{\epsilon \sim N(0,1)}[\Ical(z)],$
which can be written as:
\begin{align}
\label{sec:pre:ei}
E[\Ical(z)] &= \int_{-\infty}^{+\infty} \Ical(z) \phi(\epsilon) d \epsilon \\
&=(\mu - f^*) \Phi(\frac{\mu - f^*}{\sigma}) + \sigma \phi(\frac{\mu - f^*}{\sigma}),\nonumber
\end{align}
where $\phi, \Phi$ are the probability density function, cumulative density function of standard normal distribution, respectively.
\label{sec:bcl}
\begin{figure*}[ht]
		\centering
    \begin{tabular}{cc}
		\includegraphics[width=2.2in]{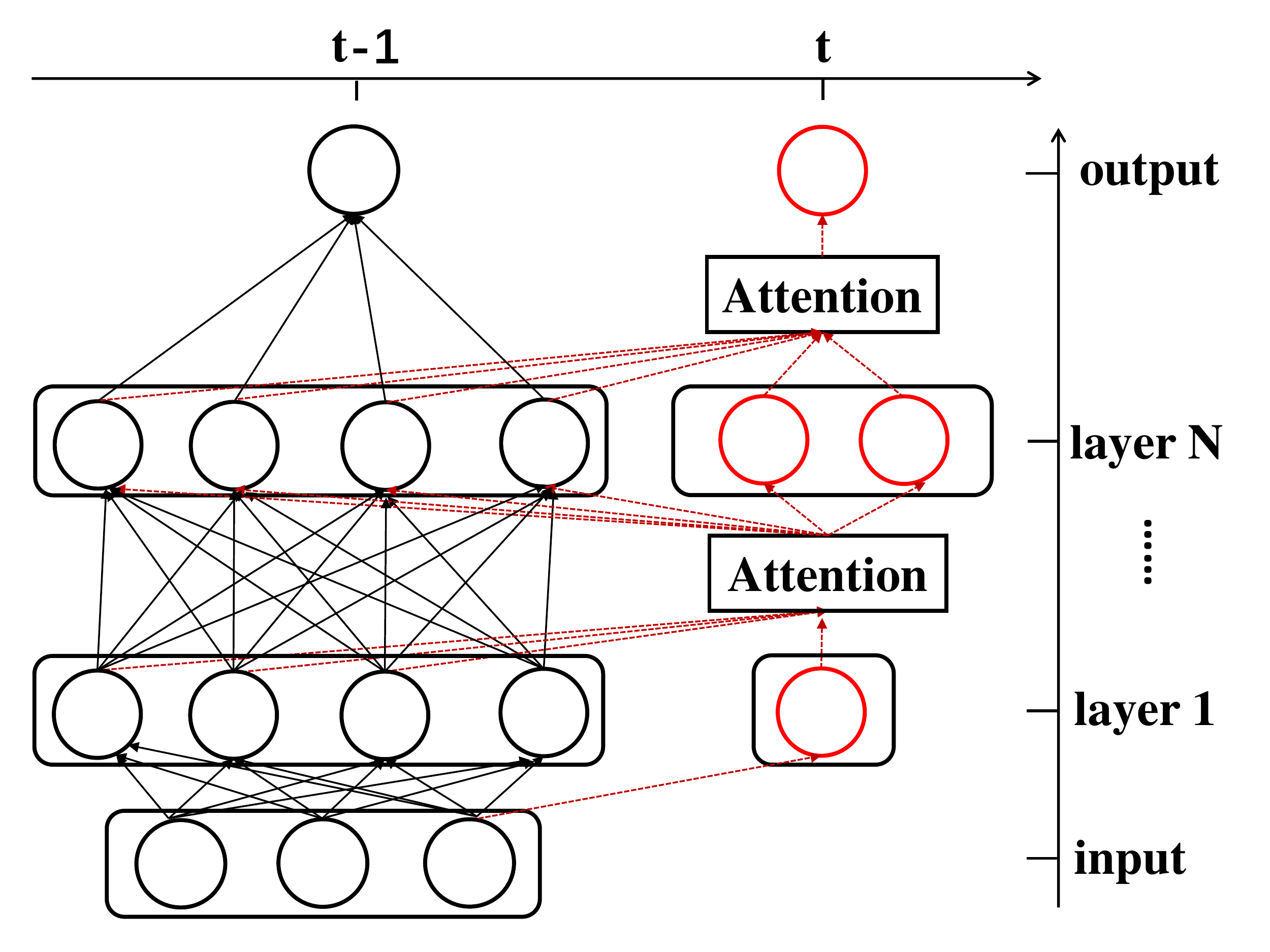} & \includegraphics[width=2.2in]{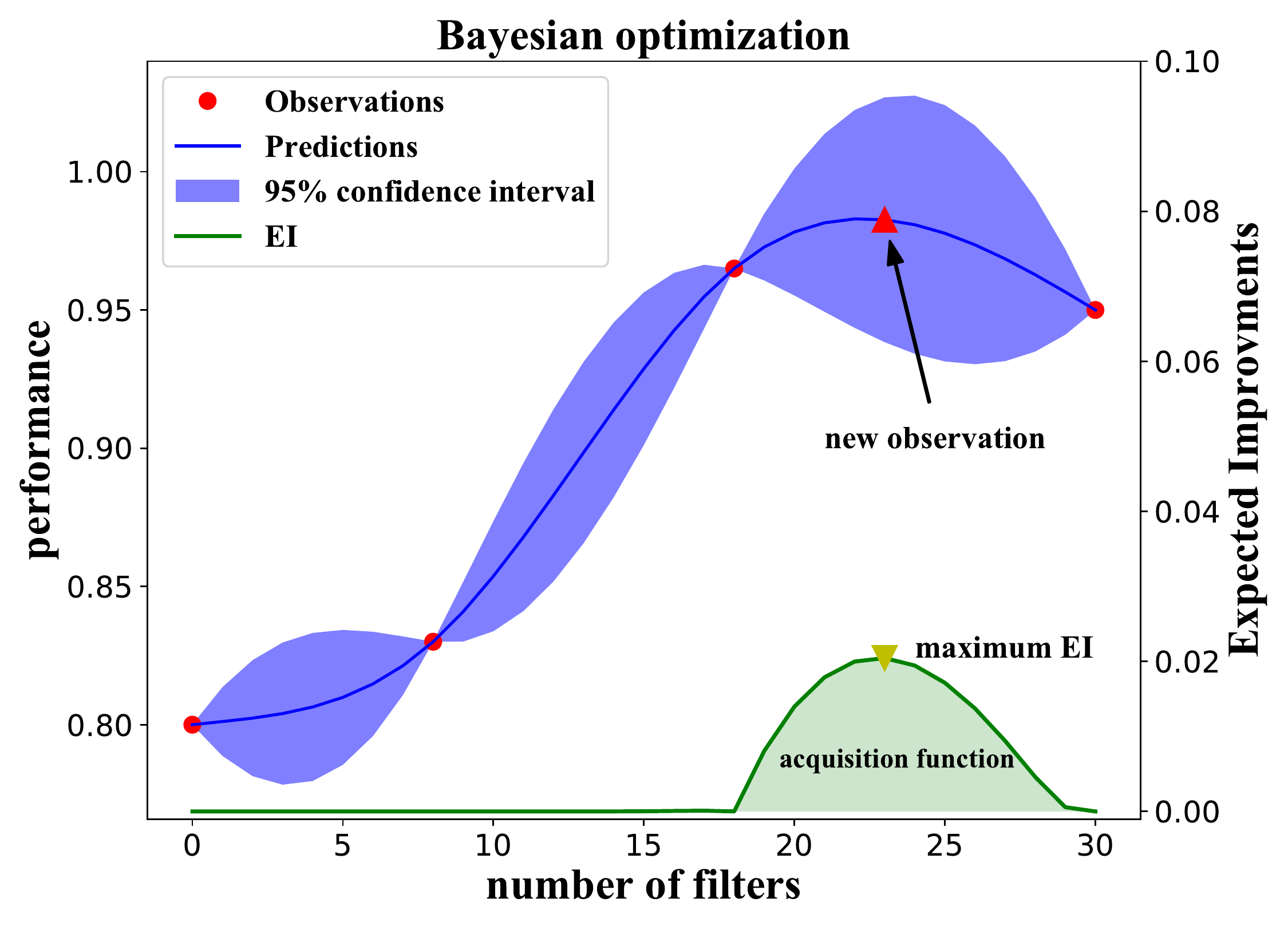} \\ 
    (a) & (b)
    \end{tabular}
         \vspace{-0.3cm}
 		 \caption{(a) BOCL expands the network adaptively when $t$-th task arrives. (b) How Bayesian optimization works in our BOCL.}
         \vspace{-0.5cm}
		\label{fig:bocl}
\end{figure*} 
\section{Our Approach: BOCL}
In this section, we elaborate on the new framework for continual learning. BOCL is composed of a Bayesian optimization procedure and a task network. We employ Bayesian optimization to decide how many filters or nodes to be added in each layer. The task network can be any network of interest for solving a particular task, such as image classification. In this paper, we  use a convolutional network (CNN) as the task network to demonstrate how BOCL adaptively expands this CNN to prevent forgetting, though our method can not only adapt to convolutional networks, but also to fully connected networks.

Figure~\ref{fig:bocl}(a) shows how BOCL expands the network and utilizes previous knowledge via attention mechanism when a new task arrives. After the learning process of task $t-1$ finishes and task $t$ arrives, we apply Bayesian optimization to decide the appropriate number of filters to be added to each layer. Figure~\ref{fig:bocl}(b) demonstrate how Bayesian optimization searches for the best architecture for a new task. 
In order to prevent semantic drift, we only train the newly added parameters. We apply attention mechanism (see Section~\ref{sec:tasknet}) for utilizing previous knowledge efficiently. 
After we have trained the model for task $t$, each newly added parameters will be memorized by the shape of every layer to prevent the caused semantic drift. During the inference time, each task only employs the parameters introduced in stage $t$,  and neglects parameters introduced in the later tasks. 
In the following, we will describe each component of our method in more details.

\subsection{Bayesian Optimization}
 The input to the ``black box'' function is a vector $\zB = (z_1, z_2, \cdots, z_n)$ (i.e., $z_n$ indicates how many filters added in $n$-th layer), and the output of the ``black box'' function $r$ is the model performance (e.g., the combination of test accuracy and model complexity shown in Eq.~\eqref{eq:reward}). First, we build a Gaussian process model based on $m$  input samples, denoted as $S = \{ (\zB^{(i)}, r^{(i)}) \}_{i=1}^m$, where $\rB = [r^{(1)}, r^{(2)},\cdots, r^{(m)}]^T$ 
 is obtained by the child network constructed according to $\zB^{(i)}$ and affected by some independent Gaussian noise $\epsilon$. $r^{(i)} = f(\zB^{(i)}) + \epsilon^{(i)}, i=1,...,m,$
where the $\epsilon^{(i)}$ are i.i.d. noise variables with independent $
\Ncal(0,\sigma^2)$ distributions. We assume a Gaussian process prior distribution over the black-box function $f(\cdot)$. $f(\cdot) \sim \Gcal \Pcal (0,k(\cdot,\cdot))$, 
where $k$ is a valid covariance, and we use Mat{\`e}rn covariance,

Now suppose $\Tcal = \{ (\zB_*^{(i)}) \}_{i=1}^{m_*}$ is a set of testing points drawn from the search space. We would like to evaluate the performance $\rB_* = [r_*^{(1)}, r_*^{(2)},\cdots, r_*^{(m)}]^T$ of points $\Tcal$ based on our Gaussian process model. For notational convenience, we define $\ZB = [(\zB^{(1)})^T,\cdots,(\zB^{(m)})^T]^T$, $\fB = [f(\zB^{(1)}), \cdots, f(\zB^{(m)})]$, $\ZB_* = [(\zB_*^{(1)})^T,\cdots,(\zB_*^{(m)})^T]^T$, $\fB_* = [f(\zB_*^{(1)}),  \cdots, f(\zB_*^{(m)})]$.

For any function $f(\cdot)$ drawn from the Gaussian process prior with covariance function $k(\cdot,\cdot)$, the  marginal distribution over any set of input points have a joint multivariate Gaussian distribution. We can represent our problem as: 
\begin{align}
\left[
 \begin{matrix}
   \rB \\
   \rB_*
  \end{matrix}
 \right]
\sim \Ncal \left(\zeroB,\left[ \begin{matrix}
   K(\ZB,\ZB) + \sigma_m^2 I&K(\ZB,\ZB_*)\\
   K(\ZB_*,\ZB)&K(\ZB_*,\ZB_*)
   \end{matrix} \right]\right),
\end{align}
where $K(\ZB,\ZB) \in R^{m \times n}$ such that $K_{ij}(\ZB,\ZB) = k(\zB^{(i)}, \zB^{(j)})$, and $K(\ZB,\ZB_*) \in R^{m \times m_*}, K(\ZB_*,\ZB) \in R^{m_* \times m}, K(\ZB_*,\ZB_*) \in R^{m_* \times m_*}$.
%
%

Now, we can obtain the distribution over the vector $\rB_*$, $\rB_* \mid (\rB,\ZB,\ZB_*)  \sim \Ncal \left( \muB^*, \SigmaB^* \right)$, 
where $\muB_* = K(\ZB_*,\ZB)(K(\ZB,\ZB) + \sigma_m^2 I)^{-1}\rB$, $\SigmaB_* = K(\ZB_*,\ZB_*)  - K(\ZB_*,\ZB)(K(\ZB,\ZB)+\sigma_m^2 I)^{-1} K(\ZB,\ZB_*)$

Based on the Gaussian process, we can directly calculate $\mu$ and $\Sigma$ of every point in our spearch space $\Zcal$ ($\mu$ and $\Sigma$ are scalar here for each point). And then we can obtain EI of each point by E.q.~\eqref{sec:pre:ei}. However it would be time consuming if the search space is huge. Therefore, we search for the best point around $\zB$ by optimization, formulated as
\begin{align}
\label{sec:method:optimization}
\max \limits_{\zB \in \Zcal} \quad &(\mu - r^*) \Phi(\frac{\mu - r^*}{\Sigma}) + \Sigma \phi(\frac{\mu - r^*}{\Sigma}) \nonumber \\
s.t. \quad &\mu = K(\zB,\ZB)(K(\ZB,\ZB) + \sigma_m^2 I)^{-1}\rB \\
 &\Sigma = - K(\zB,\ZB)(K(\ZB,\ZB)+\sigma_m^2 I)^{-1} K(\ZB,\zB) + \nonumber\\
 &\quad \quad K(\zB,\zB) ,\nonumber 
\end{align}
where $r^*$ is the best model performance observed so far. We utilize L-BFGS algorithm to solve this problem. The solution is denoted as $\zB^{(m+1)}$. We now can build a child network based on $\zB^{(m+1)}$ and evaluate its performance $r^{(m+1)}$. And then, the point $(\zB^{(m+1)},r^{(m+1)})$ will be added in the training set $S$ for next training round.


\subsection{The Task Network}
\label{sec:tasknet}
We deal with  $T$ tasks arriving in a sequential manner with training dataset $\Dcal_t=\{x_i,y_i\}_{i=1}^{N_t}$ , validation dataset $\Vcal_t=\{x_i,y_i\}_{i=1}^{M_t}$, test dataset $\Tcal_t = \{x_i,y_i\}_{i=1}^{K_t}$ at time $t$. 
For the first task, we train a basic task network that performs well enough via solving a standard supervised learning problem,
\begin{alignat}{2}
\label{init}   
\min_{W_1} L_1(W_1;\Dcal_1),
\end{alignat}
where $W_1$ are the parameters of the network. We define the well-trained parameters as $\hat{W}_t$ for task $t$. When the $t$-th task arrives, we have already known the best parameters $\hat{W}_{t-1}$ for task $t-1$. Now we utilize Bayesian optimization to decide how many filters should be added to each layer, and then we obtain an expanded child network, whose parameters to be learned are denoted as $W_t$ (including $\hat{W}_{t-1}$). 

\paragraph{Attention mechanism for utilizing previous knowledge.} In order to decrease the number of added parameters when learning new tasks, we should utilize some features learnt from previous tasks that contribute to the learning of new tasks and drop some features learnt from previous tasks that do harm to the learning of new tasks. Inspired by the attention method in image captioning task~\cite{chen2017sca}, 
we introduce  channel-wise attention (for convolution neural networks) and node-wise attention (for fully connected networks) to learn which parts of knowledge obtained from previous tasks are useful for the new tasks. This is in contrast to existing continual learning approaches (e.g., RCL, DEN, PGN) applying all the learned knowledge, which might interfere with the performance of learning new tasks. 

For channel-wise attention, if the feature maps of layer $n-1$ is $\Ucal_{n-1} \in \mathbb{R}^{W\times H \times C}$, $\Ucal_{n-1} = [u_1,u_2,\cdots,u_C]$, where $u_i \in \mathbb{R}
^{W\times H}$ reprsents the \textit{i}-th channel of the feature map, and $C$ is the total number of channels. And then each channel will be applied average pooling to obtain the channel feature $v = [v_1,v_2,\cdots,v_C], v \in \mathbb{R}^{C}$,
where scalar $v_i$ is the mean of vector $u_i$. Then $\Fcal_n$ will be calculated as follows:
\begin{align}
&\beta^{n} = sigmoid(W_s^n(relu(W_c^n v + b_c^n))+b_s^n) \\
&\Fcal_n = conv(\Fcal_{n-1} \odot \beta^n,W_{t,1}^n ),
\end{align}
where $W_{t,1}^n$ are the convolutional parameters in \textit{n}-th layer, $W_c^n$, $b_c^n$, $W_s^n$, $b_s^n$ are attention parameters in \textit{n}-th layer, $\beta^{n} \in \mathbb{R}^C$, $\odot$ is the element-wise product operation \textit{on each channel}. 

For node-wise attention, if the feature maps of fully connected layer $n-1$ is $\Fcal_{n-1} \in \mathbb{R}^{C\times 1}$, then the $\Fcal_{n}$ will be calculated as follows:
 \begin{align}
 &\beta^{n} = sigmoid(W_s^n(relu(W_c^n \Fcal_{n-1} + b_c^n))+b_s^n) \\
 &\Fcal_n = conv(\Fcal_{n-1} \odot \beta^n,W_{t,1}^n ),
 \end{align}
where $\odot$ is the element-wise product operation \textit{on each node}. 

The training procedure for the new task is as follows, keeping $\hat{W}_{t-1}$ fixed and only back-propagating the newly added parameters of $W_t \backslash \hat{W}_{t-1}$. Thus, the optimization formula for the new task is $\min_{W_t \backslash \hat{W}_{t-1}} L_t(W_t;\Dcal_t)$.

We use stochastic gradient descent to learn the newly added parameters with weight decay, and $\eta$ is the learning rate,
\begin{alignat}{2}
\label{update}
& W_t \backslash \hat{W}_{t-1} \longleftarrow W_t \backslash \hat{W}_{t-1} - \eta \nabla_{W_t \backslash \hat{W}_{t-1}}(L_t).
\end{alignat}

We will train the expanded child network until the required number of epochs. And then the child network will be tested on the validation dataset $\Vcal_t$ and the corresponding accuracy $A_t$ will be returned. 
The parameters of the expanded network achieving the best performance will be the optimal ones for task $t$, and we store them for later tasks.
\subsection{Evaluation of Points}
\label{sec:per}
The point selected by Bayesian optimization is labeled as $\zB = (z_1, z_2, \cdots, z_n)$ (i.e., $z_i$ indicates how many filters added in layer $i$). We then design a new architecture for a child network based on $\zB$, which will be trained in a new task. At convergence, this child network will achieve an accuracy $A_t$ on a validation dataset and the model complexity $B_t$.  The performance of a network is defined as follows:
\begin{equation}
\label{eq:reward}
r_t = A_t(\zB) + B_t(\zB), 
\end{equation}
where $B_t(\zB) = -\sum\limits_{i=1}^n z_i \alpha_i$, $\alpha_i = \alpha {P_i}/(\sum\limits_{j=1}^n P_j)$, $P_j$ is the number of parameters added in \textit{j}-th layer, and $\alpha$ is a hyper-parameter to balance  the prediction performance and model complexity. Compared with the performance design in ~\cite{xu2018reinforced} that utilized the same $\alpha$ in every layer, we employ dynamic $\alpha$ for different layers. Our performance design can peform better and strike a better balance between model accuracy and model complexity. For example, suppose a convolutional layer $C_1 = conv(3,3,3,12)$ and a fully-connected layer $F_1 \in \mathbb{R}^{(400\times 120)}$, add one filter in layer $C_1$ will bring less parameters than add one node in layer $F_1$. 

\subsection{Training Procedures}
\label{training}
Bayesian optimization typically needs several initialization points. If these initialization points perform poorly, a large number of  trials are required to obtain satisfactory performance. 
 To reduce  the search time, we propose to employ the previous search experience of Bayesian optimization to warmstart the search for the new task based on the similarity of task difficulty (described by certain task meta-feature). 
In our method, we build an episodic memory $\mathbb{M}$, and after finishing $(t-1)$-th task, we  augment the meta-feature $M_{t-1}$ and the best point $\zB^{(t-1)}$ to the memory:  $\mathbb{M}^{(t-1)} = \{(M_j, \zB^{(j)}) \}_{j=1}^{t-1}$. When a new task arrives, we  choose several points from $\mathbb{M}^{(t-1)}$ based on the $L_1$ distance of meta-features to warmstart our Bayesian optimization. More details are elaborated in the following.

When the $t$-th task comes, we propose the meta-feature of $t$-th task as $M_t = A_t^1 - A_t^2$, where the test accuracy $A_t^1$ is obtained from training the the task on the base network (e.g. LeNet for CIFAR-100, fully connected networks for MNIST, details can be saw in section ~\ref{sec:exp}). 
And we train the $t$-th task based on the parameters $\hat{W}_{t-1,1}$ and newly added attention parameters. Only the attention parameters will be finetuned,  while $\hat{W}_{t-1,1}$ will be fixed. We can obtain the test accuracy $A_t^2$. Intuitively, 
$M_t$ can measure the accuracy gap that we can increase by expanding our base network. If the accuracy gap $M_t$ is big, we may need add more filters/nodes; otherwise,  only a few even no filters/nodes are required to add. After that, $m=3$ points $\{ \zB^{(i)} \}_{i=1}^m$  will be selected from $\mathbb{M}^{(t-1)}$ (if the number of points in $\mathbb{M}^{(t-1)}$ is less than $m$, the rest of points will be randomly selected from our search space) based on the $L_1$ distance between  meta-feature $M_t$ and each element in $\{ M_i\}_{i=1}^{(t-1)}$. Then  the child network will be trained to obtain corresponding performance, $\{ (r^{(i)}) \}_{i=1}^m$ according to Eq.~\eqref{eq:reward}. Now we can utilize $ S= \{ (\zB^{(i)}, r^{(i)}) \}_{i=1}^m$ to warmstart an initial GP model.

Based on the built GP, we can select the next point $\zB^{m+1}$  via Eq.~\eqref{sec:method:optimization}. This point will be utilized as a configuration to build a new child network (the rounded value of $\zB^{m+1}$ is used when constructing a child network since the number of filters is an integer), after training which the corresponding performance $r^{m+1}$ could be obtained. And then $\{ (\zB^{m+1}, r^{m+1}) \}$ will be added into $S$ for next training round.

We will repeat the above process until we reach the maximum trial number $N$ or the performance ceases to increase in $H$ trials. Finally, we can obtain a series of performance and corresponding points (i.e. architecture configurations), and then choose the point that has the best performance as our final result.  We summarize our BOCL approach in Algorithm~\ref{alg:bcl}, and its subroutine for network expansion is described in Algorithm~\ref{alg:expanding}.

\begin{algorithm}[tb]
 \caption{BOCL for Continual Learning}
  \label{alg:bcl}
  \begin{algorithmic}[1]
    \STATE {\bfseries Input:} A sequence of datasets $\Dcal=\{ \Dcal_1, \Dcal_2,\dots,\Dcal_T \}$ 
    \STATE {\bfseries Output:} $\hat{W}_T$
    \FOR{$t=1,...,T $}
    \IF {$t=1$}
    \STATE Train the base network using ~\eqref{init} on the first datasest $\Dcal_1$ and obtain $\hat{W}_1$.
    \ELSE
    \STATE Expand  the network by Algorithm~\ref{alg:expanding}, and obtain $\hat{W}_t$.
    \ENDIF
    \ENDFOR
  \end{algorithmic}
\end{algorithm}

\begin{algorithm}[tb]
  \caption{Routine for Network Expansion}
  \label{alg:expanding}
  \begin{algorithmic}[1]
    \STATE {\bfseries Input:} Current dataset $\Dcal_t$; previous parameter $\hat{W}_{(t-1)}$; episodic memory $\mathbb{M}^{t-1}$;  number of trials, $N$; number of trials if the performance doesn't increase, $H$.
    \STATE {\bfseries Output:} Network parameter $\hat{W}_t$
    \STATE Obtain initialization points $S = \{ (\zB^{(i)},{} r^{(i)}) \}_{i=1}^m$ based on $\mathbb{M}^{(t-1)}$ to warmstart GP (details in Section ~\ref{training});
    \FOR{$i=1,\dots,N $}
    \STATE Select the next best point $\zB_t^i$  shown in Eq.~\eqref{sec:method:optimization}.
    \STATE Build an expanded child network according to $\zB_t^i$; 
    \STATE Train the expanded network using Eq.~\eqref{update} to obtain $W_t^{(i)}$ and  performance $r_t^i$ evaluated by Eq.~\eqref{eq:reward}.
    \STATE Break the loop if the performance doesn't increase in $H$ trials.
    \STATE Add $(\zB_t^i,r_t^i)$ to training data $S$. $S \leftarrow \{ S, (\zB_t^i,r_t^i) \}\nonumber$
    \STATE Adjusting the GP using training points $S$;
    \ENDFOR
    \STATE Return the best network parameter configuration, $\hat{W}_t  = \argmax_{W_t^{(i)}} r_t(W_t^{(i)})$.
  \end{algorithmic}
\end{algorithm}

\begin{figure*}[ht]
		\centering
    \begin{tabular}{ccc}
		\includegraphics[width=2in]{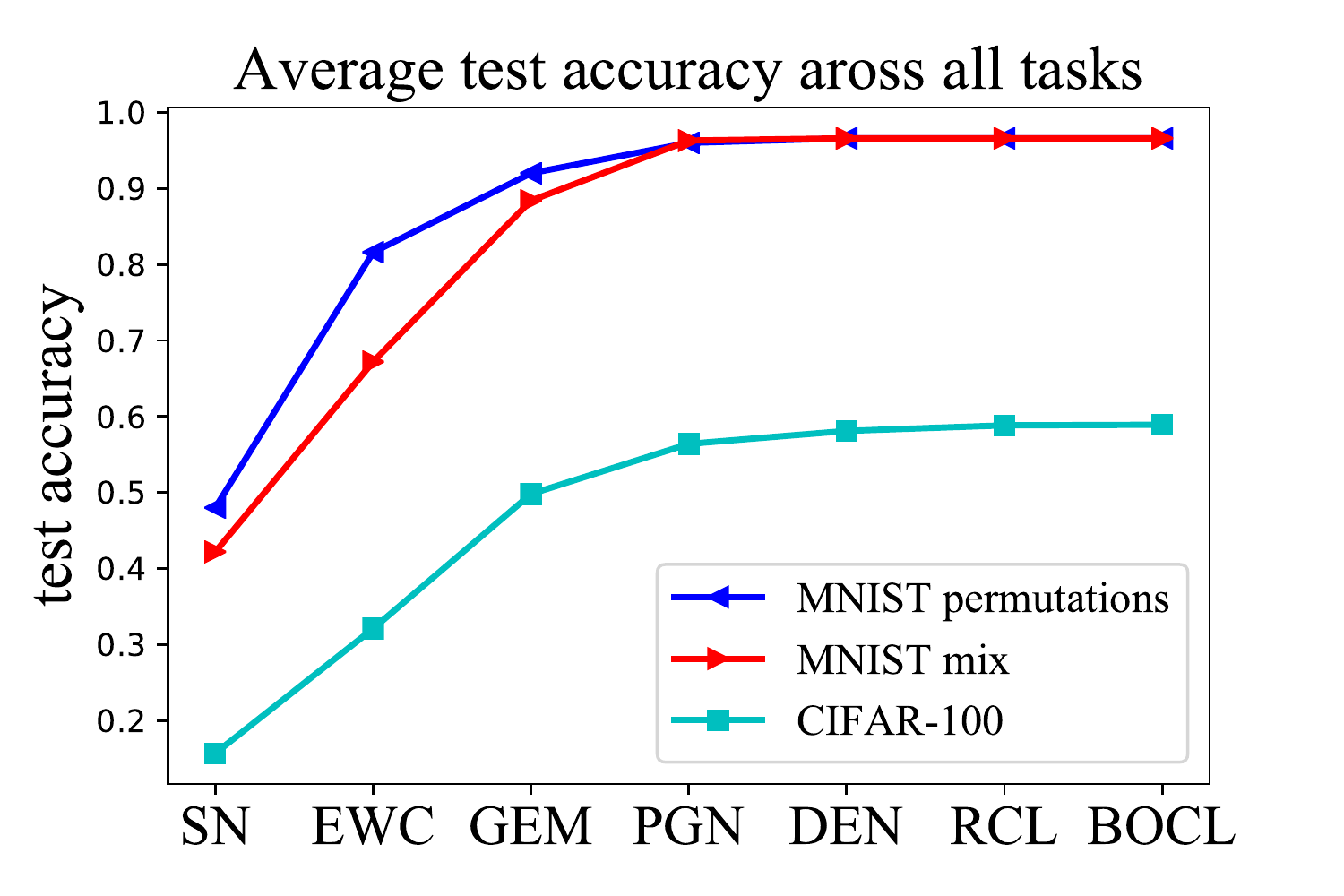} & \includegraphics[width=2in]{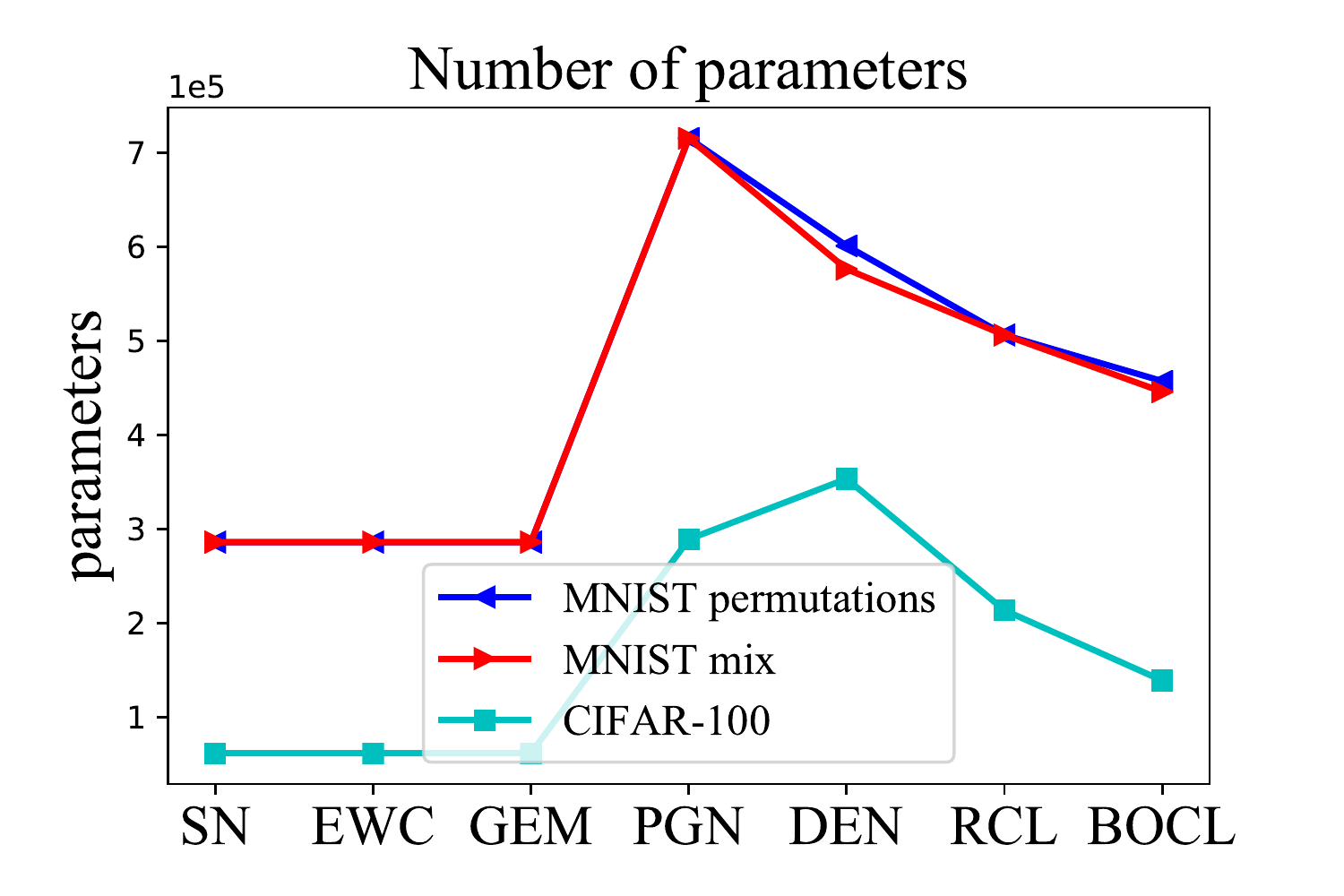} & \includegraphics[width=2in]{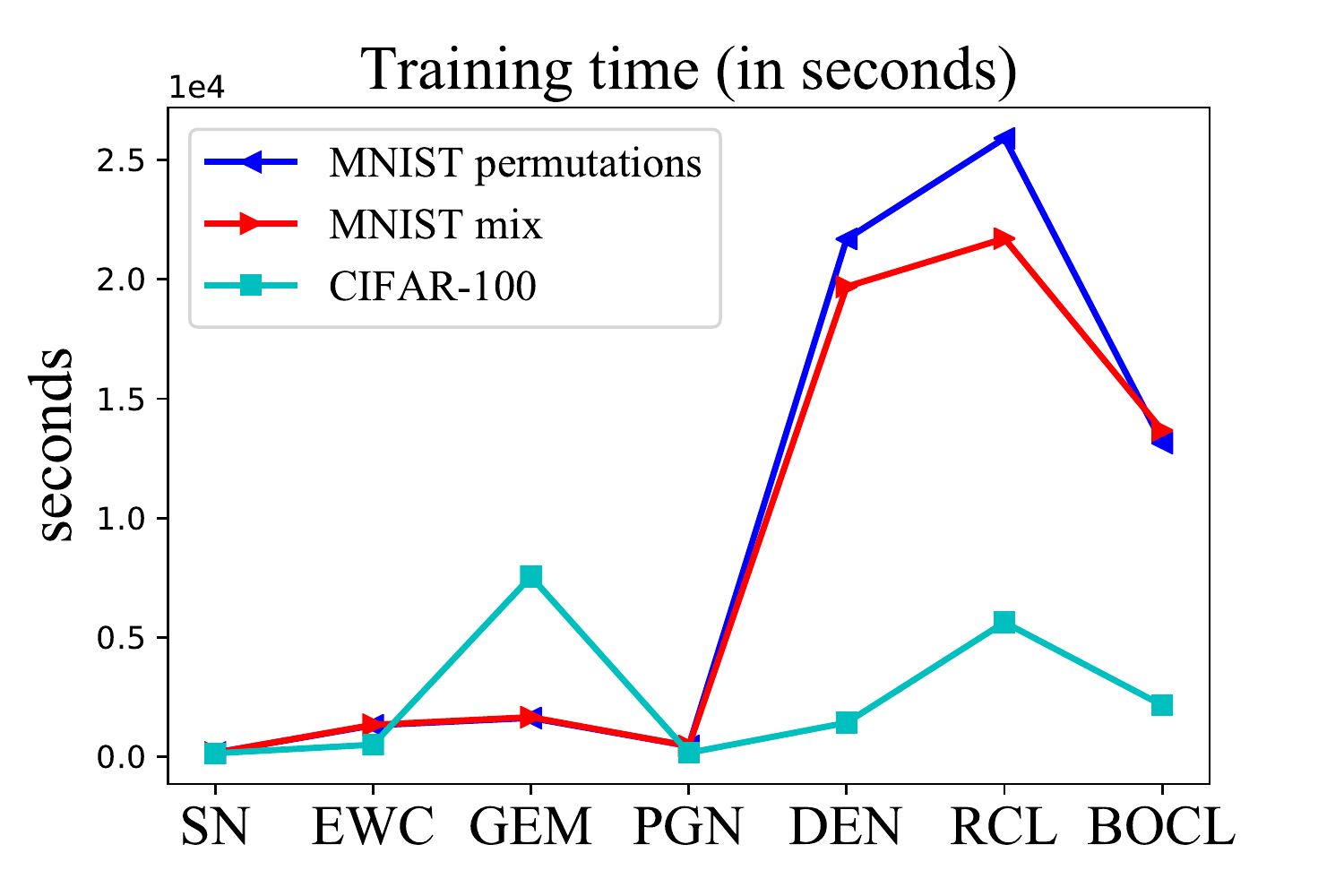}\\ 
    (a) & (b) &(c)
    \end{tabular}
         \vspace{-0.3cm}
 		 \caption{(a) Average test accuracy for all datasets. (b) Number of parameters after all tasks finished. (c) Training time for different methods.}
         \vspace{-0.5cm}
		\label{experiments}
\end{figure*} 


\section{Experiments}
\label{sec:exp}
We perform a variety of experiments to assess the performance of BOCL for continual learning. We will report the accuracy and the model complexity achieved by our BCL and the state-of-the-art baselines. We implemented all the experiments in Tensorfolw on GPU Tesla K80. 
\paragraph{Datasets}
(1) \textbf{MNIST Permutations}~\cite{kirkpatrick1}. Ten variants of the MNIST data, where each variant is constructed by different random permutation of MNIST pixels; (2) \textbf{MNIST Mix}~\cite{xu2018reinforced}. Ten variants of MNIST data $\Dcal=\{ \Dcal_1, \dots,\Dcal_{10} \}$, where $\{ \Dcal_1, \Dcal_3, 
\Dcal_5, \Dcal_7, \Dcal_9 \}$ are MNIST Permutations, $\{ \Dcal_2, \Dcal_4, 
\Dcal_6, \Dcal_8, \Dcal_{10} \}$ are MNIST Rotations \cite{GradientEpisodicMemory}. MNIST Rotations are a variant of MNIST where each contains digits rotated by a fixed angle between 0 and 180 degrees.  (3) \textbf{Incremental CIFAR-100}~\cite{icart}. A variants of the CIFAR-100. Different from the original CIFAR-100, each task introduces a new set of classes. For the total number of tasks $T$, each new task contains digits from a subset of $100/T$ classes. 

For all of the above datasets, we set the number of tasks to be learned as $T = 10$. For the MNIST datasets, each task contains 60000 training examples and 1000 test examples from 10 different classes. For the CIFAR-100 datasets, each task contains 5000 train examples and 1000 examples from 10 different classes. The model observes the tasks one by one, and once the task had been observed, the task will not be observed later during the training.

\paragraph{Baselines}
(1) \textbf{SN}, a single network trained across all tasks; (2) \textbf{EWC}, avoid catastrophic forgetting by regularization with elastic weight consolidation~\cite{kirkpatrick1}; (3) \textbf{GEM}, gradient episodic memory for continual learning~\cite{GradientEpisodicMemory}; (4) \textbf{PGN}, progressive neural network~\cite{rusu1}; (5) \textbf{DEN}, dynamically expandable network by regularization~\cite{yoon1}; (6) \textbf{RCL}, dynamically expandable network by reinforcement learning~\cite{xu2018reinforced}.

\paragraph{Base network settings}
(1) Fully connected networks (three-layer network with 784-312-128-10 neurons) for  MNIST Permutations and MNIST Mix datasets; (2) LeNet~\cite{lenet1} is used for Incremental CIFAR-100. 
\begin{figure*}[t]
  \centering
    \begin{minipage}[b]{0.78\textwidth}
      \centering
      \includegraphics[width=1\textwidth]{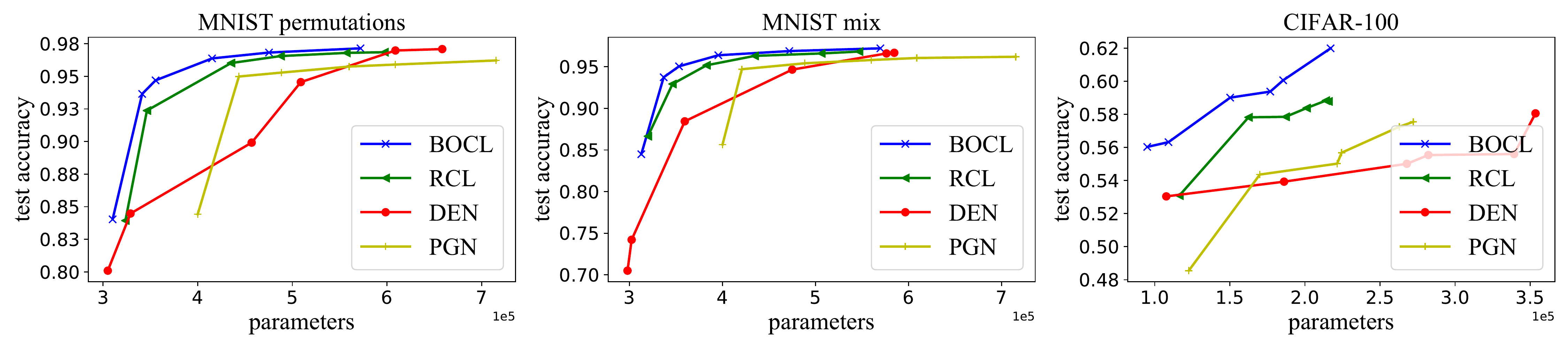}
      \vspace{-0.8cm}
      \caption{Average test accuracy \emph{v.s.} model complexity for RCL, DEN and PGN.}
      \vspace{0.15cm}
      \label{fig:accvspara}
    \end{minipage}
    \begin{minipage}[b]{0.78\textwidth}
      \centering
    \includegraphics[width=1\textwidth]{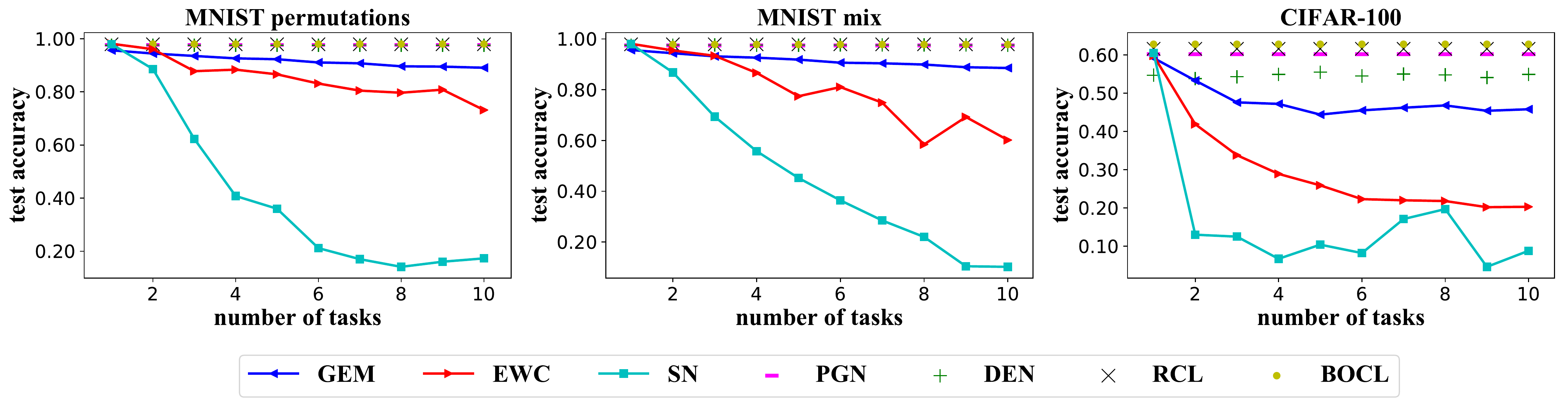}
    \vspace{-0.8cm}
    \caption{Test accuracy on the first task as more tasks are learned.}
     \vspace{0.15cm}
     \label{fig:forgetting}
    \end{minipage}
    \begin{minipage}[b]{0.78\textwidth}
     \centering
   \includegraphics[width=1\textwidth]{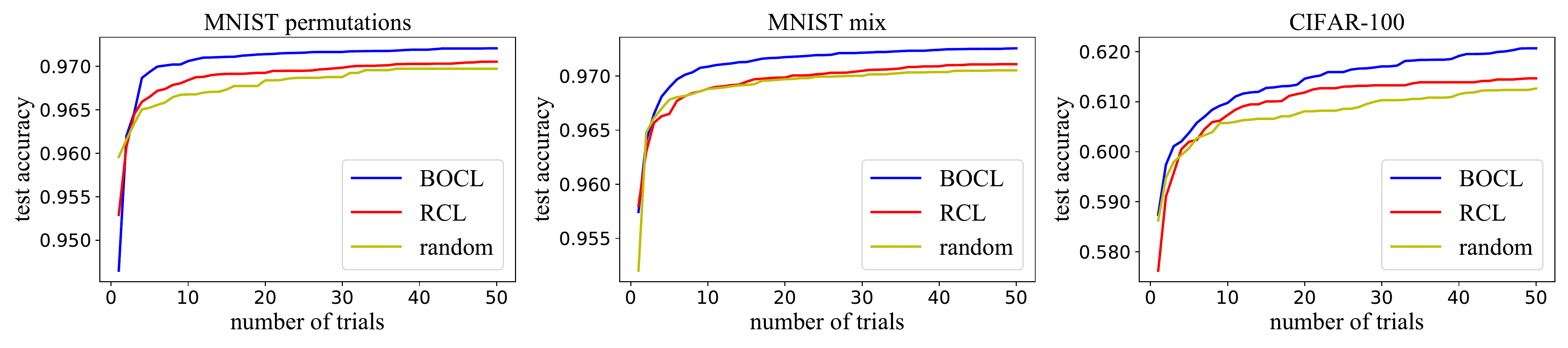}
   \vspace{-0.8cm}
   \caption{Average test accuracy on all tasks as the number of trials increases.}
   \label{fig:trials}
    \end{minipage}
\end{figure*}
\subsection{Results}
We evaluate each compared approach by considering  average test accuracy, model complexity and training time on all the tasks. We evaluate model complexity via the number of the number of model parameters after finishing all the subtasks. The results can be found  in Figure~\ref{experiments}. 

As shown in Figure~\ref{experiments}(a) and Figure~\ref{experiments}(b), we can conclude that methods with fixed-size networks such as SN, EWC and GEM own terrible test accuracy and low model complexity. In sharp contrast, methods with expandable networks can fit new tasks well and prevent forgetting previous tasks better. This demonstrates that dynamically expanding networks can indeed increase model accuracy significantly.

\textbf{Comparison with PGN, DEN and RCL.} 
We tune the hyperparameters in our BOCL to make the average accuracy is on par with RCL, DEN and PGN or slightly better than them. As shown in Figure~\ref{experiments}(b),  we can conclude that BOCL achieves significant reduction on the number of parameters (also shown in Table ~\ref{table2}), especially on CIFAR-100.
\begin{table}[htbp]
    \vspace{-0.2cm}
  \setlength{\abovecaptionskip}{-2pt}
  \caption{Reduction on the number of parameters.}
  \centering
  \begin{tabular}{lccc}
    \toprule
    Methods &PGN&DEN&RCL \\
    \hline
    MNIST permutations&$36\%$&$24\%$&$10\%$\\
    MNIST mix&$38\%$&$23\%$&$12\%$\\
    CIFAR-100&$52\%$&$60\%$&$35\%$\\
    \bottomrule
  \end{tabular}
  \setlength{\belowcaptionskip}{-2pt}
  \vspace{-0.3cm}
  \label{table2}
\end{table} 

As for training time in Figure~\ref{experiments}(c), compared with RCL, it can be easily observed that our proposed method BOCL achieves 49\%, 37\%,  62\% reduction on training time on MNIST permutations, MNIST mix, CIFAR-100, respectively. Besides, our BOCL achieves 39\%, 31\% reduction on training time on MNIST permutations, MNIST mix respectively compared with DEN. However, our BOCL takes about 51\% more training time than DEN on CIFAR-100.

To further compare the differences between these methods, we vary some hyperparameters (e.g., weight for regularization terms in DEN, $\alpha$ in RCL and BOCL) in each method and train until convergence. These hyperparameters will affect the model complexity and test accuracy. Then we can obtain how test accuracy changes with respect to the number of parameters, as shown in Figure~\ref{fig:accvspara}. We can clearly see that BOCL can obtain higher test accuracy with the same number of parameters, and fewer parameters with the same test accuracy. This demonstrates that our method BCL achieves the best performance even with different hyperparameters.

\textbf{Evaluating the forgetting behavior.} 
One major point we concentrate on in continual learning is that models should perform well in previous tasks while learning more new tasks. We evaluate the forgetting behavior through measuring the test accuracy on the first task while learning the increasing number of tasks. Figure~\ref{fig:forgetting} shows the evolution of the test accuracy on the first task as more tasks are learned. We can clearly see that BOCL, RCL and PGN prevent catastrophic forgetting totally while the other methods cause different degrees of catastrophic forgetting. We can observe that DEN still cause catastrophic forgetting since it finetunes previous parameters while learning new tasks.

\textbf{The importance of attention mechanism.} 
In order to utilize previous knowledge more efficiently while learning new tasks, we apply attention mechanism to our model. We will demonstrate that the attention we proposed can indeed contribute to the improvement of model performance.

In our experiment setup, hyperparameters are the same except the attention mechanism (with attention v.s. without attention). We run each experiment for four times. We find that the model without attention achieves 0.1\% less accuracy on MNIST permutations, 0.16\% less accuracy on MNIST mix and 2.2\% less accuracy on CIFAR-100 compared with the model with attention.

We can see that attention mechanism performs better on CIFAR-100 than on MNIST, the reason for which is that models without attention have obtained a pretty good performance on MNIST. Therefore the space for improvement on MNIST is very limited. In contrast, since model without attention performs not so well on CIFAR-100, attention mechanism can indeed contribute to the model performance by a large margin. 

\textbf{Comparing with random search and reinforcement learning.} We compare different searching methods in our proposal (Bayesian optimization v.s. random search, reinforcement learning). In every experiment setup, hyper-parameters are the same except the searching method. Each experiment is run for four times. As shown in Figure~\ref{fig:trials}, we can conclude that Bayesian optimization costs the smallest number of trials with the same accuracy.


\section{Conclusion}
\label{sec:con}
We have introduced BOCL, a novel model for continual learning that prevents catastrophic forgetting completely and utilizes previous knowledge better.  Our method searches the best network architecture for new tasks via Bayesian optimization. Besides, we apply attention mechanism in our method to utilize previous knowledge more effectively. We validate our method's competitive performance on different datasets. 

\bibliographystyle{named}
\bibliography{ijcai19}

\end{document}